\documentclass[letterpaper, 10 pt, journal, twoside]{ieeetran} 

\IEEEoverridecommandlockouts                              

\usepackage{graphicx}
\usepackage{hyperref}
\usepackage{tikz}
\usepackage{makecell}
\usepackage{diagbox}
\usepackage{orcidlink}
\usepackage[ruled,vlined]{algorithm2e}
\SetKw{Continue}{continue}
\SetKw{Break}{break}
\LinesNumbered
\SetArgSty{textup}
\hypersetup{
    colorlinks=true,
}
\usepackage[font=footnotesize,labelsep=period]{caption}[2022/02/20]
\usepackage{subcaption}
\usepackage{amsfonts}
\usepackage{amsmath}
\usepackage[thinlines]{easytable}
\usepackage{hhline}
\usepackage{stackengine}
\newcommand\xrowht[2][0]{\addstackgap[.5\dimexpr#2\relax]{\vphantom{#1}}}
\usepackage{amsthm} 
\usepackage{cite}
\usepackage{booktabs}

\usepackage{duckuments}

\definecolor{myRed}{RGB}{255, 29, 37}
\definecolor{myOrange}{RGB}{255, 147, 30}
\definecolor{myGreen}{RGB}{122, 201, 67}
\definecolor{myBlue}{RGB}{63, 169, 245}

\title{\LARGE \bf
Time-Optimal Path Planning in a Constant Wind for Uncrewed Aerial Vehicles using Dubins Set Classification\\
} 

\begin{document}
\author{Brady Moon$^{*1}$, Sagar Sachdev$^{*2}$, Junbin Yuan$^{2}$, and Sebastian Scherer$^{1}$
\thanks{*The first two authors contributed equally to this work.}
\thanks{Manuscript received: June, 20, 2023; Revised September, 19, 2023; Accepted October, 27, 2023. This paper was recommended for publication by Editor Pauline Pounds upon evaluation of the Associate Editor and Reviewers' comments.
This work is supported by the Office of Naval Research (Grant N00014-21-1-2110). This material is based upon work supported by the National Science Foundation Graduate Research Fellowship under Grant No. DGE1745016. \textit{(Corresponding author: Brady Moon.)}} 
\thanks{$^{1}$Authors are with the Robotics Institute, School of Computer Science at Carnegie Mellon University, Pittsburgh, PA, USA
{\tt\footnotesize  \{bradym, basti\}@andrew.cmu.edu}}%
\thanks{$^{2}$Authors are with the Mechanical Engineering Department at Carnegie Mellon University, Pittsburgh, PA, USA {\tt\footnotesize \{sagarsac, junbiny\}@andrew.cmu.edu}}%
}

\markboth{IEEE Robotics and Automation Letters. Preprint Version. Accepted Oct, 2023}
{Moon \MakeLowercase{\textit{et al.}}: Time-Optimal Path Planning in Wind for UAVs} 

\maketitle

\begin{abstract}
Time-optimal path planning in high winds for a turning-rate constrained UAV is a challenging problem to solve and is important for deployment and field operations. Previous works have used trochoidal path segments comprising straight and maximum-rate turn segments, as optimal extremal paths in uniform wind conditions. Current methods iterate over all candidate trochoidal trajectory types and select the one that is time-optimal; however, this exhaustive search can be computationally slow. In this paper, we introduce a method to decrease the computation time. This is achieved by reducing the number of candidate trochoidal trajectory types by framing the problem in the air-relative frame and bounding the solution within a subset of candidate trajectories. Our method reduces overall computation by 37.4\% compared to pre-existing methods in Bang-Straight-Bang trajectories, freeing up computation for other onboard processes and can lead to significant total computational reductions when solving many trochoidal paths. When used within the framework of a global path planner, faster state expansions help find solutions faster or compute higher-quality paths. We also release our open-source codebase as a C++ package.  
\href{https://bradymoon.com/trochoids}{[Website]}\footnote{ \href{https://bradymoon.com/trochoids}{Website: https://bradymoon.com/trochoids}}
\href{https://github.com/castacks/trochoids}{[Codebase]}\footnote{ \href{https://github.com/castacks/trochoids}{Codebase: https://github.com/castacks/trochoids}}
\href{https://youtu.be/qOU5gI7JshI}{[Video]}\footnote{ \href{https://youtu.be/qOU5gI7JshI}{Video: https://youtu.be/qOU5gI7JshI}}


\end{abstract}

\section{Introduction}


\IEEEPARstart{D}{eploying} robotic platforms in the field often necessitates taking into account additional disturbances and challenges introduced by the environment. For example, the perception could be affected by changing lighting conditions, glare, and precipitation, while the path planning and controls are affected by disturbances and changing environmental factors such as wind \cite{Kulkarni2020}. Ignoring these factors in path planning can lead to less robust results, suboptimal paths, infeasible paths, or even the loss of the vehicle due to collisions.

\begin{figure}[th]
    \centering
    \vspace{-2mm}
    \includegraphics[trim={2.4cm 2.3cm 7.5cm 4.1cm},clip,width=.49\textwidth]{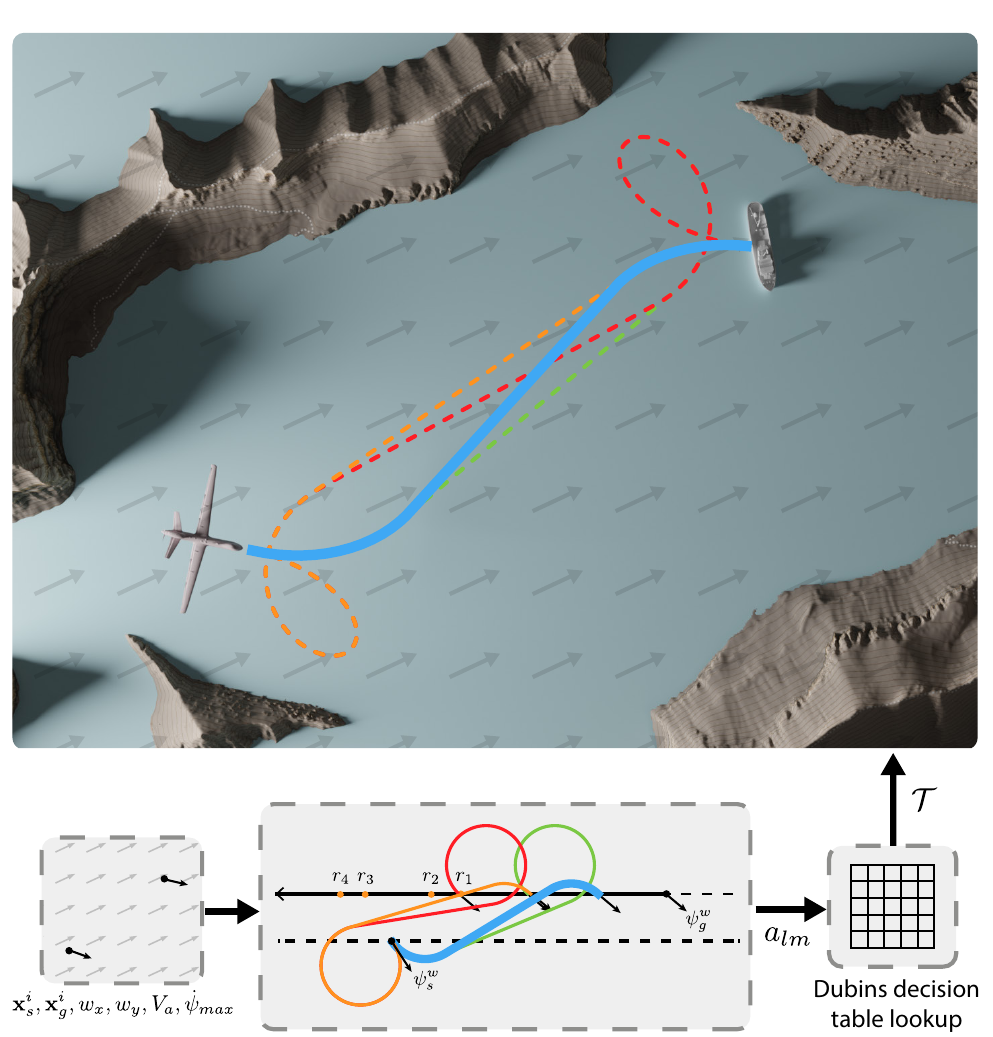} 
    \vspace{.2mm}
    \caption{An overview of our method, with an example problem. The input conditions are used to find the quadrant pair for the solution to then look up the set of candidate trajectories in the Dubins decision table. In this case, the only candidate trajectory is LSR, thus reducing the problem to solving only one out of the four potential trajectories. The bottom middle section shows the four extremal paths transformed into $\mathcal{F}^w$, where the goal state is moving in the negative $x$ direction with the same speed as the wind. It is clear to see that the optimal LSR path is the first to intersect with the goal state. The upper image shows the four potential extremal paths in $\mathcal{F}^i$, with the blue LSR path being optimal. }
    \vspace{-2mm}
    \label{fig:full_overview}
\end{figure}

Connecting two points on a plane with a time-optimal path using a curvature-constrained motion model has been shown by \cite{dubinsPaper} and proven by \cite{Johnson1974} to be a path made up of maximum curvature turns and straight segments called a Dubins path. This solution is simple, quick to compute, and though not curvature continuous, has been applied in a variety of robotic applications such as Autonomous Underwater Vehicles (AUV) \cite{wang2017} and Uncrewed Aerial Vehicles (UAV) \cite{Owen2015}. However, the Dubins path does not consider the effect of wind or ocean currents on the optimal path.

This paper describes a framework for quickly finding time-optimal paths for curvature-constrained vehicles in a uniform flow field, such as UAVs in wind fields, AUVs in deep ocean currents, or autonomous surface vehicles (ASVs) in surface currents. We consider a kinematic model for a UAV flying at a constant altitude and constant air-relative speeds in a steady uniform wind.

Previous works solved this problem through a computationally inefficient process of computing all possible trajectory types and selecting the minimum-time path from the set of computed trajectories. The most recent approach to this problem by \cite{techy2009} was able to find a closed-form solution to quickly compute the solution for two of the four possible Bang-Straight-Bang (BSB) trajectories; however, the method still has to perform a costly global root-finding search using the Newton-Raphson method for the remaining two trajectories. Our method reduces the set of possible optimal BSB trajectories before having to find them, thus reducing overall computation time. To the best of our knowledge, this work is the state-of-the-art solution for computing time-optimal trajectories in a uniform flow field while constrained by a dynamics model. 
Our key contributions are the following:
\begin{itemize}
    \item An approach that formulates planning curvature-constrained paths in wind fields into a domain that allows for using Dubins set classification.
    \item A novel mechanism to divide the solutions space into segments where each has a reduced set of possible optimal BSB trajectories.
    \item A method to find which segment contains the optimal solution, reducing the set of possible optimal BSB trajectories to fewer than four and decreasing the overall computation time by 37.4\%. 
    \item Our open-source codebase as a package for others to use, benchmark, test, and improve.
\end{itemize}

This paper is organized as follows: Section \ref{sec:related} specifies the related works, Section \ref{sec:problem} outlines the problem definition, Section \ref{sec:method} explains our methodology in depth, Section \ref{sec:results} presents our results, and Section \ref{sec:conclusion} gives the conclusion and future work.

\section{Related Work}\label{sec:related}
Several approaches have been used for path-planning in uniform wind conditions given dynamic curvature constraints, namely, vehicle velocity and turning radius. 

In the absence of wind conditions, the work by \cite{shkel2001} presents an approach for planning time-optimal paths using Dubins Curves, as were first formulated by Dubins in \cite{dubinsPaper}. However, instead of having to iterate over all the feasible trajectories (four in the case where the distance between the start and goal position is relatively large), the authors classify the Dubins set and present a decision table wherein they can identify the optimal path from the set based on the start and goal configurations of the UAV. This reduces the number of candidate trajectories to below four for each cell of the decision table. This table produces sixteen combinations of possible quadrants, which can be represented in the form $a_{lm}$ where $l$ is the quadrant number of the initial orientation and $m$ is the quadrant number of the final orientation. A switching function then chooses a single trajectory from the feasible candidate trajectory in the decision table cells. A modified version of the decision table can be found in Table \ref{table:new_matrix} within the methodology section.

\begin{figure}
    \centering
    \includegraphics[trim={0cm 0cm 1.5cm 1.5cm},clip,width=\columnwidth]{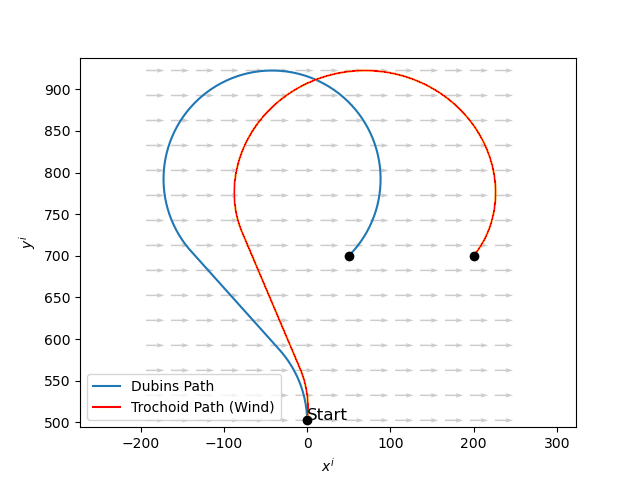} 
    \caption{An example optimal trochoidal LSR paths and its equivalent Dubins path in the wind frame. }
        \label{fig:Dubins_Frame_vs_Wind_frame}
\end{figure}

Some of the previous works have used a frame transformation to align the axis with the wind. \cite{mcgee2005} and \cite{mcgee2007} use this frame transformation to augment the problem into a moving-goal problem, wherein the velocity of the goal point is equal and opposite to the velocity of the wind. Finding the intersection point is referred to as the ``Rendezvous Problem." The problem states that there are six admissible Dubins Paths, namely RSR, LSL, RSL, LSR, RLR, and LRL, wherein ``R" represents a maximum turn-rate right turn, ``L" a maximum turn-rate left turn and ``S" a Straight Line. The solutions to this problem can either be solvable or unsolvable. If the problem is solvable, the paths and their lengths can be computed for all six admissible paths using a standard numerical root-finding approach such as the Newton-Raphson method. The minimum-time path is chosen as the optimal path. Fig.~\ref{fig:Dubins_Frame_vs_Wind_frame} shows a Dubins Path augmented into a trochoidal path when a wind vector is applied. 

Another optimal control approach was also proposed by \cite{bakolas2010} wherein it tries to solve the Markov-Dubins problem and the Zermelo's navigation problem (which was also formulated in \cite{mcgee2007}) in order to find the time-optimal path to a target while avoiding obstacles and respecting the kinematic constraints of the vehicle. A key difference between this method in \cite{bakolas2010} and that of \cite{mcgee2007} was that it proposes using backward dynamic programming to compute the optimal control policy and value function for vehicles with Dubins' type kinematics operating in the presence of wind rather than using the Pontryagin's maximum principle.

The work by \cite{techy2009} also uses the same frame-alignment of \cite{mcgee2005} and \cite{mcgee2007}. However, a key distinction is that they provide a closed-form solution to two of the six admissible trajectory types: RSR and LSL. To find the remaining four paths, it proposes using a standard numerical global root-finding approach such as Newton-Raphson. The optimal path is then found by computing the lengths of the above six paths and choosing the path with the minimum time. In contrast, our work introduces a method that circumvents the necessity to compute paths for all trajectory types by identifying the subset of trajectories that contains the optimal path and only computing paths for that subset.



\section{Problem Definition}\label{sec:problem}

Let the system dynamics be defined by the equations
\begin{align*}
    \dot{x}^i(t) &= V_a \cos\left({\psi^i (t)}\right) + w_x \\
    \dot{y}^i(t) &= V_a \sin\left({\psi^i (t)}\right) + w_y \\
    \dot{\psi}^i(t) &= u(t)
\end{align*}
where the state is given by $\mathbf{x}= [ x, y, \psi ]^\top$, $i$ denotes the inertial frame $\mathcal{F}^i$, $t$ is time, $V_a$ is the vehicle airspeed, $w_x$ is the wind speed in the $x$ axis, $w_y$ is the wind speed in the $y$ axis, and $u$ is the control input. 

The objective is to find the control inputs that yield the minimum time path between an initial state $\mathbf{x}_s^i  = [ x_s^i, y_s^i, \psi_s^i  ]^\top$ and final state $\mathbf{x}_g^i = [ x_g^i, y_g^i, \psi_g^i ]^\top$ given the control constraint $\| u \| < \dot{\psi}_{max}$. As was shown in \cite{mcgee2005,techy2009}, the solution to this optimization problem is one of the six extremal paths.



\section{Methodology}\label{sec:method}



On a high level, our trochoidal path planner first formulates the problem as the goal state moving at a uniform velocity in the opposite direction of the wind flow, similar to \cite{mcgee2005}. We then calculate transition points where the initial or final quadrants change and determine between which transition points the solution lies. Because the quadrants of the initial and final points are constant between the transition points, we can leverage the decision table presented in \cite{shkel2001} to predict the type of trajectory we expect to be optimal. We then employ the approach presented in \cite{techy2009}, wherein an analytical method is used to find the solution to RSR and LSL trajectories, and a global root-finding approach is used to find the solution to RSL and LSR trajectories given in the decision table. This methodology improves overall computational efficiency and speed. An overview of our method is outlined in Algorithm \ref{alg:ours}.


The first step is to align the $x$ axis with the wind, transforming the inertial frame $\mathcal{F}^i$ to the wind frame $\mathcal{F}^w$. The wind magnitude can be calculated at $V_w = \sqrt{v_x^2 + v_y^2}$, and the wind angle, $\psi_w^i$, can thus be computed as
\begin{align}  
    \psi_w^i = \tan^{-1} \left(\frac{v_y}{v_x}\right).
    \label{equ:one}
\end{align}
Let $\psi_s^w$ and $\psi_g^w$ be the start and goal state angles defined as
\begin{align}
    \psi_s^w &= \psi_s^i - \psi_w^i \label{equ:two}\\
    \psi_g^w &= \psi_g^i - \psi_w^i \label{equ:three}.
\end{align}
The $x$ and $y$ elements of the start and goal states are transformed from $\mathcal{F}^i$ to $\mathcal{F}^w$ with the rotational matrix
\begin{equation}
R_i^w = 
    \begin{bmatrix}
    \cos (\psi_w^i) &  \sin (\psi_w^i) \\
    -\sin (\psi_w^i) & \cos (\psi_w^i) 
    \end{bmatrix}.
    \label{equ:four}
\end{equation}
\begin{algorithm}[th]
\SetInd{0.4em}{0.8em}
Set $\mathbf{x}_s^w, \mathbf{x}_g^w $ using (\ref{equ:one}-\ref{equ:four})\;
$\Gamma \leftarrow \emptyset$\;
$\delta \theta_\alpha \leftarrow CalcDeltaTheta(\psi_s^w)$\;
$\delta \theta_\beta \leftarrow CalcDeltaTheta(\psi_g^w)$\;
$\Gamma \leftarrow \{CalcTransitionPoints(\psi_s^w,\delta \theta_\alpha)\}$\;
$\Gamma \leftarrow \Gamma \cup \{CalcTransitionPoints(\psi_g^w,\delta \theta_\beta)\}$\;
\For{$r_n \in \Gamma$}{
 
\If{$CheckSegment4R(r_n)$}{
$\mathcal{T} \leftarrow$ \{RSR,RSL,LSL,LSR\}\;
\Break\;
}
$l_p \leftarrow DubinsPath(r_n)$\;
$t_{start} = l_p/V_a$;
$t_{goal} = r_n/V_w$\;
\If{$t_{start} \leq t_{goal}$}{
Compute $\theta_q$\ using (\ref{equ:thetaq})\;
\Break\;
}
}
\If{$\theta_q$ is empty}{
Compute $\theta_q$ as angle between $r_N$ and $\pi$\;
}
$\alpha_q = \psi_s^w - \theta_q$; $\beta_q = \psi_g^w - \theta_q$\;
$a_{lm} \leftarrow FindQuadrants(\alpha_q, \beta_q)$\;
$\mathcal{T} \leftarrow DecisionTable(a_{lm})$\;
\For{each candidate trochoidal trajectory in $\mathcal{T}$}{
Solve for $u$ using (\ref{equ:solveTroch1}-\ref{equ:solveTroch3})\;
\If{$PathTime(u) < PathTime(u_{best}$)}{
$u_{best} \leftarrow u\; $
}
}
\Return $u_{best}$ 
\caption{Trochoid($\mathbf{x}_s^i$, $\mathbf{x}_g^i$, $w_x$, $w_y$, $V_a$, $\dot{\psi}_{max}$)
\label{alg:ours}}
\end{algorithm}
The problem can now be viewed as finding the minimum time intersection point of the vehicle and goals state, with the wind being removed from the vehicle system dynamics and viewing the goals state as moving in the negative $x$ direction with the velocity $V_w$. The updated system dynamics of the vehicle in the frame $\mathcal{F}^w$ would be 
\begin{align*}
    \dot{x}^w(t) &= V_a \cos\left({\psi^w (t)}\right) \\
    \dot{y}^w(t) &= V_a \sin\left({\psi^w (t)}\right) \\
    \dot{\psi}^w(t) &= u(t)
\end{align*}
and the goal state position at time $t$ would be

\begin{align*}
    \mathbf{x}_g^w(t) &= 
    \begin{bmatrix}
        R_i^w & 0 \\
        0 & 1
    \end{bmatrix}
    \mathbf{x}_g^i
    -
    \begin{bmatrix}
        tV_w \\
        0 \\
        \psi_w^i
    \end{bmatrix}.
\end{align*}

\begin{figure}[th]
    \centering
    \vspace{2mm}
    \includegraphics[trim={0cm 0cm 0cm 0cm},clip,width=.48\textwidth]{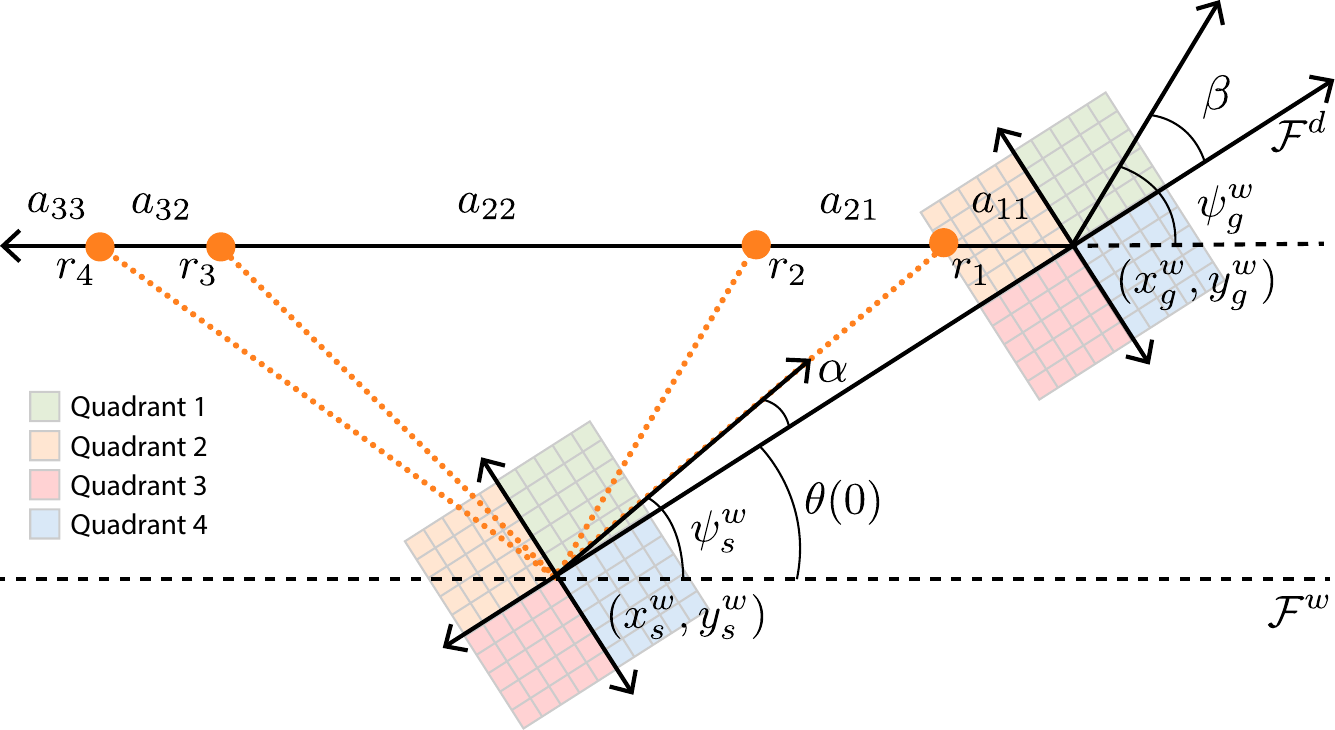} 
    \caption{~~An example problem setup with variable definition shown. In this example, there are four transition points, which is the maximum. Depending on $V_a$, $w_x$, $w_y$, and $\dot{\psi}_{max}$, the solution would lie within one of the quadrant pair sections.}
    \label{fig:decisionLIne}
\end{figure}
With the problem formulated in this manner, connecting any two \textit{static} points in $\mathcal{F}^w$ can be treated as a Dubins problem, as there is no wind acting on the vehicle. A Dubins path can be solved with just the distance between the two states $d$, and the orientation of the start and goal states relative to the axis connecting the two states. We define this axis-aligned frame as $\mathcal{F}^d$ shown in Fig.~\ref{fig:decisionLIne}. To find the Dubins path between the static start state and a goal state parameterized by a given $t$, the angles $\psi_s^w$ and $\psi_g^w$ are transformed to the Dubins frame $\mathcal{F}^d$ by first computing the angle, $\theta$, between the two points with 
\begin{equation*}
\theta(t) = \tan^{-1}\left(\frac{y_g^w (t) - y_s^w}{x_g^w (t) - x_s^w}\right).
\end{equation*} 
The initial and the final angles in $\mathcal{F}^d$, $\alpha$ and $\beta$, can be computed as
\begin{align*}
\alpha(t) &= \psi_s^w - \theta(t) \\
\beta(t) &= \psi_g^w - \theta(t).
\end{align*}
With $\alpha$, $\beta$, and $d$, the Dubins path between the states can be found. If $d>4R_0$, where $R_0$ is the turning radius, we can directly look up the optimal extremal path type based on the quadrants of $\alpha$ and $\beta$ \cite{shkel2001}. After solving for the optimal path, the time of path traversal for the vehicle from the $\mathbf{x}_s^w$ to $\mathbf{x}_g^w(t)$ is $t_{start}= {l_p}/{V_a}$, where $l_p$ is the length of the path. The time for the initial goal state $\mathbf{x}_g^w(0)$ to reach $\mathbf{x}_g^w(t)$ at the speed of $V_w$ is $t_{goal}=t$. If $t_{start}$ is equal to $t_{goal}$, then this would be the rendezvous time and state.

However, as the goal point moves in the direction opposite to the wind, the $\theta$ changes along with both $\alpha(t)$ and $\beta(t)$, meaning we cannot directly look up the optimal extremal path type. We address this by first dividing the goal state trajectory into sections based on the points along the trajectory where $\alpha(t)$ or $\beta(t)$ transitions into new quadrants. Each section defines a period where the quadrants of $\alpha(t)$ and $\beta(t)$ are constant. If we can find which section of the path contains the solution, we can constrain the solution space based on the quadrants of $\alpha(t)$ and $\beta(t)$.

\textit{Proposition 1:} If the solution lies between the transition points $r_n$ and $r_{n-1}$, then the set of candidate optimal trajectories $\mathcal{T}$ is the cell in the Dubins decision table associated with any $\alpha$ or $\beta$ angle derived from a point between $r_n$ and $r_{n-1}$.

\textit{Proof:} Let the quadrant of $\alpha$ be $l$, the quadrant of $\beta$ be $m$, and the quadrant pair $a_{lm}$ denote the cell of row $l$ and column $m$ in the Dubins decision table. Any transition point $r_n$ denotes a change of $l$ or $m$. Therefore between $r_n$ and $r_{n-1}$, $l$ and $m$ stay constant. Therefore $a_{lm}$ is the same for all $\alpha$ and $\beta$ between the transition points. Any of such values can be used to find $a_{lm}$ and the corresponding set of candidate optimal trajectories $\mathcal{T}$ from the cell in the decision table. \hfill\qedsymbol

We define the point where $\alpha(t)$ or $\beta(t)$ transitions to a new quadrant as $r_n$, where $n \in N$, $N$ is the total number of transition points, and $\Gamma$ is the ordered set of transition points.

The first quadrant transition for $\alpha(t)$ is at $\theta_{\alpha 1} =\delta \theta_\alpha + \theta(0)$ where $\delta \theta_\alpha$ is the rotation for $\alpha(t)$ to transition to a new quadrant and is positive when $y_g^w$ is positive and negative when $y_g^w$ is negative. This is due to $\theta(t)$ rotating in the positive direction when the goal state $\mathbf{x}_g^w(0)$ is above the start state $\mathbf{x}_s^w$ and rotating in the negative direction when the goal state is below the start state.

If the rotation from $\theta(0)$ to $\theta_{\alpha 1}$ doesn't cross the $x$ axis, then the transition point can be computed by 
\begin{align}
    r_{\alpha 1} = x_g^w - x_s^w - \frac{y_g^w-y_s^w}{\tan{ \left(  \theta_{\alpha 1}\right)}}
    \label{equ:rn}
\end{align}
which is the distance from $\mathbf{x}_g^w(0)$ where $\mathbf{x}_g^w(t)$ intersects with the line from $\mathbf{x}_s^w$ in the $\theta_{\alpha 1}$ direction. There is no intersection point if $\theta_{\alpha 1}$ crosses the $x$ axis and is directed away from the goal state trajectory.

The following transition point angle $\theta_{\alpha 2}$ is $\pi/2$ from $\theta_{\alpha 1}$ in the same direction as $\delta \theta_\alpha $. Just as before, if $\theta_{\alpha 2}$ does not cross the $x$ axis in the rotation from $\theta(0)$, then the next transition point is calculated with (\ref{equ:rn}) using $\theta_{\alpha 2}$. This process is then repeated for $\beta(t)$ to find its transition points and both are added to the ordered set $\Gamma$. $N$ can be at most $4$, because a rotation of $\pi/2$ from either of the latter points would lead to an angle pointing away from the goal state trajectory. An example configuration can be seen in Fig.~\ref{fig:decisionLIne}.

The transition points are then iterated over from smallest to largest to check if the solution lies before that point. First, all points in the segment between the transition point and the previous one are checked for the $d > 4R_0$ condition. If the check is satisfied, the time it takes to go from the goal state to the transition point $r_n$ is computed by $t_{goal} = {r_n}/{V_w}$. Similarly, the time for the vehicle to travel from the start state to the transition point is also computed by finding the Dubins path between the points and then solving for the time with $t_{start} = {l_p}/{V_a}$, where $l_p$ is the path length. If $t_{start} < t_{goal}$, it indicates that the solution lies between the transition points, and the quadrants for that specific segment now need to be determined. 

To find the quadrants for the segment, we compute the angle of the midpoint between the transition points with  
\begin{align}
    \theta_q = \tan^{-1}\bigg(\frac{y_g-y_s}{x_g - x_s - \frac{r_n + r_{n-1}}{2}}\bigg) \label{equ:thetaq}
\end{align}
and $r_0 = 0$ for the case when finding the quadrant of the first segment.
Using the midpoint rather than the transition point ensures that the lookup of the quadrants does not lie on a boundary. 

If the solution does not lie between any of the transition points ($t_{start} > t_{goal}$ for all transition points), then the solution lies beyond the last transition point. If that happens, rather than computing the angle of the midpoint between transition points, we directly compute the $\theta_q$ as halfway between the last transition point and $\pi$.


\begin{table}[t]
    \centering
    \vspace{3mm}
    \resizebox{1\columnwidth}{!}{
    \renewcommand{\arraystretch}{1.5}
    \begin{tabular}{|c||*{4}{c|}}
        \hline
        \backslashbox{\bf$Q_s$}{\bf $Q_g$}
        &\makebox[5em]{\bf 1}&\makebox[5em]{\bf 2}&\makebox[5em]{\bf 3} &\makebox[5em]{\bf 4} \\
        \hhline{|=||=|=|=|=|} \xrowht{25pt}
        \bf 1 & RSL & \makecell[c]{RSR \\ RSL \\ LSR} & \makecell[c]{RSR \\ LSR} & \makecell[c]{LSR \\ RSL \\ RSR} \\
        \hline \xrowht{25pt}
        \bf 2 & \makecell[c]{LSL \\ RSL \\ LSR} & \makecell[c]{LSL \\ RSL \\ RSR} & RSR & \makecell[c]{RSR \\ RSL} \\
        \hline \xrowht{25pt}
        \bf 3 & \makecell[c]{LSL \\ LSR} & LSL & \makecell[c]{RSR \\ LSR \\ LSL} & \makecell[c]{RSR \\ LSR \\ RSL}\\
        \hline \xrowht{25pt}
        \bf 4 & \makecell[c]{RSL \\ LSR \\ LSL} & \makecell[c]{LSL \\ RSL} & \makecell[c]{LSL \\ LSR \\ RSL} & LSR \\
        \hline
    \end{tabular}
    }
    \caption{~~The above Dubins decision table is adapted from \cite{shkel2001} with corrections to \{$a_{12}, a_{21}, a_{34}$, $a_{43}$\}. ``R" represents a right-turn segment, ``S" represents a straight segment, and ``L" represents a left-turn segment. $Q_s$ is the quadrant of the start state, and $Q_g$ is the quadrant of the goal state. }
    \label{table:new_matrix}
\end{table}

After finding $\theta_q$, the associated initial and final angles $\alpha_q$ and the $\beta_q$ are found using the following equation:
\begin{align*}
    \alpha_q &= \psi_s^w - \theta_q\\
    \beta_q &= \psi_g^w - \theta_q .
\end{align*}
We can then find the quadrant $l$ of $\alpha_q$ and the quadrant $m$ of $\beta_q$, creating the pair $a_{lm}$. We check the corresponding block in the Dubins decision table and let $\mathcal{T}$ represent the set of candidate trajectories within that block. Because $\alpha(t)$ and $\beta(t)$ are not constant between transition points, the switching functions are not used to further narrow down the candidate trajectories, and the more conservative long path case of $4R_0$ is used rather than the precise definition given in the Dubins set classification.


During the implementation and testing of the Dubins decision table based on \cite{shkel2001}, inconsistencies between the decision table and result were seen. This error was validated by computing and comparing all six admissible paths in the Dubins set. We found the long path condition in Proposition 5 of their paper was incorrect and that the decision table sometimes did not give the actual shortest path for quadrant pairs $a_{lm} \in \{ a_{12}, a_{21}, a_{34}, a_{43} \}$. These four cases are equivalent with symmetry, and we describe the issue in detail within the Appendix. A recent work \cite{Lim2023dubins} also identified these same issues and proposes the modifications to fix the issues. A corrected version of the Dubins decision table is also provided in Table~\ref{table:new_matrix}, where an additional path type had to be added to each of the four equivalency groups with errors.

After finding $\mathcal{T}$, we solve the trochoid paths within the set using the method described in \cite{techy2009} and choose the one with minimum time. However, we now no longer have to iterate over all candidate solutions in most cases. As seen in Table \ref{table:new_matrix}, all blocks have less than four path types, with some having only one. Only when the goal state might be within $4R_0$ do all path types have to be checked.


For paths of type RSR or LSL, the solutions can be found analytically using the values
\begin{align}
    x_{s_0} &= x_s^w - \frac{V_a}{\delta_1 \omega\sin{(\psi_s^w})}\label{equ:solveTroch1}\\
    y_{s_0} &= y_s^w + \frac{V_a}{\delta_1 \omega\cos{(\psi_s^w)}}\\
    x_{g_0} &= x_g^w - \frac{V_a}{\delta_2 \omega\sin{\left(\delta_2 \omega t_{2\pi}\right) + \psi_g^w}} - V_w t_{2\pi}\\
    y_{g_0} &= y_g^w + \frac{V_a}{\delta_2 \omega\cos{\left(\delta_2\omega t_{2\pi}\right) + \psi_g^w}}.
\end{align}
The point of exit of the first bank $t_1$ and the point of entry to the second bank $t_2$ for LSL and RSR trajectories can be found with
\begin{align}
    \gamma &= \tan^{-1}\bigg(\frac{y_{g_0} - y_{s_0}}{x_{g_0} - x_{s_0} + V_w\frac{\psi_s^w - \psi_g^w + 2k\pi}{\delta_2\omega}}\bigg)\\
    t_1 &= \frac{t_{2\pi}}{\delta_1 2\pi}\left(\sin^{-1}\left({\frac{V_w}{V_a}\sin{(\gamma)}}\right) + \gamma - \psi_s^w\right)\\
    t_2 &= t_1 + \frac{\psi_s^w - \psi_g^w + 2k\pi}{\delta_2 \omega}, \nonumber \\
    & \qquad \qquad \qquad \qquad \qquad k \in \{-3, -2, -1, 0, 1, 2\} \label{equ:t2}
\end{align}
where $\delta_1 \in \{-1, 1\}$ and $\delta_2 \in \{-1, 1\}$ result in a left or right turn for the first and second bank, $\omega = |\dot{\psi}_{max}|$, $t_{2\pi}={2\pi}/{\omega}$, and $\psi_s^w - \psi_g^w$ is modulo $2\pi$.

If the trajectory is of type RSL or LSR, it can be found using a numerical global root-finding technique such as the Newton-Raphson solver to find the roots of the equation
\begin{align}
    f(t_1) = E\cos(\delta_1 \omega t_1 &+ \psi_s^w) \nonumber\\ 
   &+ F\sin({\delta_1\omega t_1 + \psi_s^w})-G 
\end{align}
where
\begin{align}
    E &= V_a \left(V_w\frac{\delta_1 - \delta_2}{\delta_1\delta_2\omega} - (y_{g_0} - y_{s_0})\right)\\
    F &= V_a \bigg(x_{g_0} - x_{s_0} + V_w\bigg(t_1\left(\frac{\delta_1}{\delta2}-1\right)\nonumber\\ & \qquad\qquad\qquad\qquad\qquad +\frac{\psi_s^w - \psi_g^w + 2k\pi}{\delta_2 \omega}\bigg)\bigg)\\
    G &= V_w(y_{g_0} - y_{s_0}) + \frac{V_a^2(\delta_2 - \delta_1)}{\delta_1 \delta_2 \omega} \label{equ:solveTroch2}
\end{align}
and then using $t_1$ to find $t_2$ using (\ref{equ:t2}). Because there are multiple roots, each has to be checked to find the valid and optimal trajectory.

\begin{figure*}[th]
     \centering
     \vspace{1mm}
     \begin{subfigure}[b]{0.32\textwidth}
         \centering
         \includegraphics[trim={0cm 0cm 1.2cm 1cm},clip,width=\textwidth]{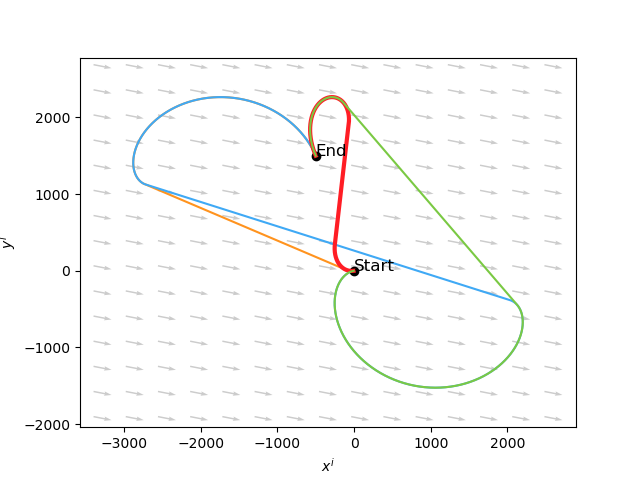}
         \label{fig:y equals x}
     \end{subfigure}
     \hfill
     \begin{subfigure}[b]{0.32\textwidth}
         \centering
         \includegraphics[trim={0cm 0cm 1.2cm 1cm},clip,width=\textwidth]{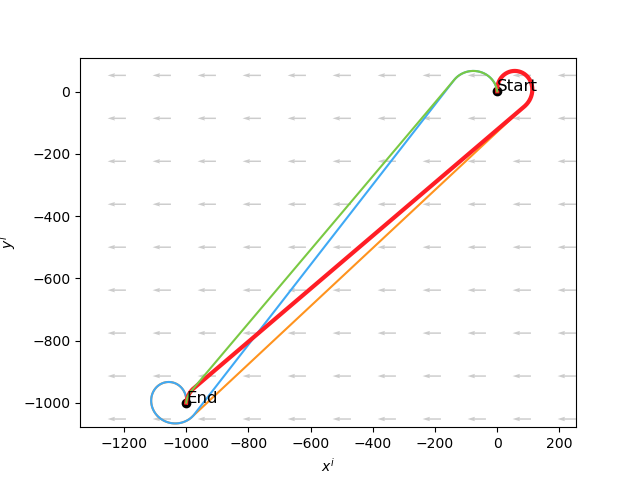}
         \label{fig:three sin x}
     \end{subfigure}
     \hfill
     \begin{subfigure}[b]{0.32\textwidth}
         \centering
         \includegraphics[trim={0cm 0cm 1.2cm 1cm},clip,width=\textwidth]{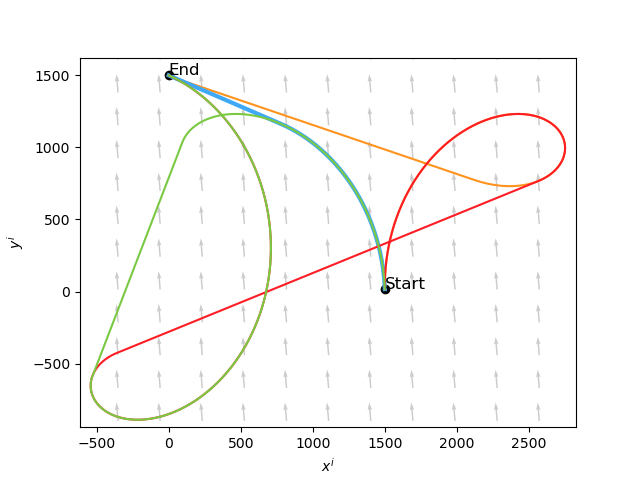}
         \label{fig:five over x}
     \end{subfigure}
     \begin{subfigure}[b]{0.32\textwidth}
         \centering
         \includegraphics[trim={0cm 0cm 1.2cm 1cm},clip,width=\textwidth]{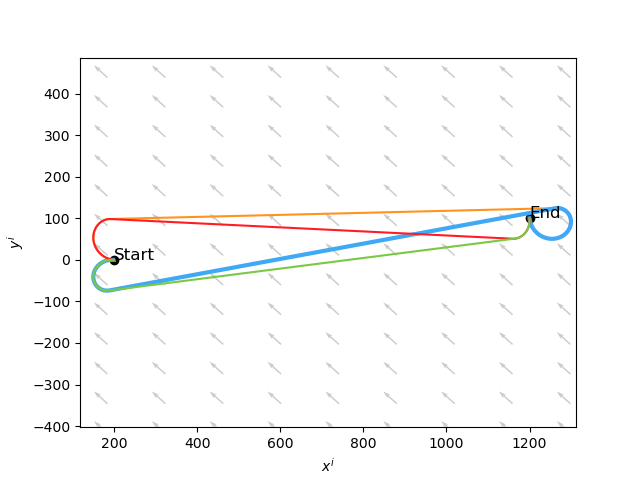}
         \label{fig:y equals x2}
     \end{subfigure}
     \hfill
     \begin{subfigure}[b]{0.32\textwidth}
         \centering
         \includegraphics[trim={0cm 0cm 1.2cm 1cm},clip,width=\textwidth]{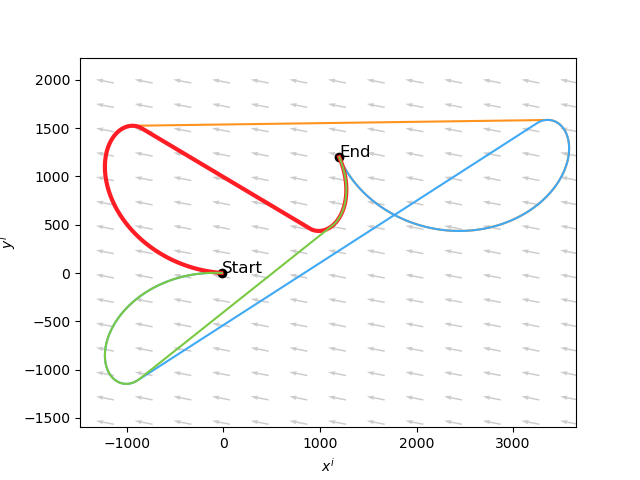}
         \label{fig:three sin x2}
     \end{subfigure}
     \hfill
     \begin{subfigure}[b]{0.32\textwidth}
         \centering
         \includegraphics[trim={0cm 0cm 1.2cm 1cm},clip,width=\textwidth]{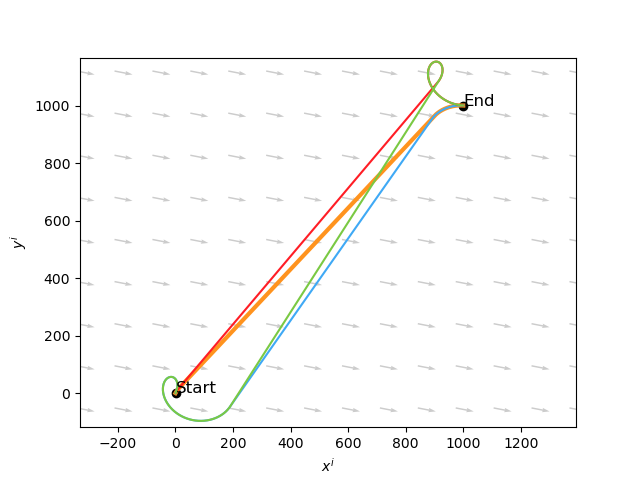}
         \label{fig:five over x2}
     \end{subfigure}
        \caption{Randomly sampled candidate BSB trajectories for randomly selected initial and final states in a uniform wind-vector field. \textcolor{myRed}{RSL} is shown as red, \textcolor{myOrange}{RSR} as orange, \textcolor{myGreen}{LSL} as green, and \textcolor{myBlue}{LSR} as blue. The bolded path in each subfigure is the optimal one.}
        \label{fig:three graphs}
\end{figure*}

After finding $t_1$ and $t_2$, the trochoidal paths can be constructed using
\begin{align}
    x_{s}^w(t) &= \frac{V_a}{\delta_1\omega}\sin\left({\delta_1\omega t + \psi_s^w}\right) + V_wt + x_{s_0}\\
    y_{s}^w(t) &= \frac{V_a}{\delta_1 \omega}\cos\left({\delta_1 \omega t + \psi_s^w}\right) + y_{s_0}\\
    x_{g}^w(t) &= \frac{V_a}{\delta_2\omega} \sin\left({\delta_2 \omega t + \psi_g^w}\right) + V_wt + x_{g_0}\\
    y_{g}^w(t) &= \frac{-V_a}{\delta_2\omega} \cos\left({\delta_2 \omega t + \psi_g^w}\right) + y_{g_0}
    \label{equ:solveTroch3}
\end{align}
where $t \in [0,t_1]$ for the first two equations to construct the first banking segment, and $t \in [t_2, t_{2\pi}]$ for the last two equations to construct the second banking segment.
For derivations of (\ref{equ:solveTroch1}-\ref{equ:solveTroch3}) and more in-depth explanations, see \cite{techy2009}.

\begin{table}[th]
    \centering
    \vspace{3mm}
    \begin{tabular}{cc}
        \toprule
        \textbf{Path Type} & \textbf{Distribution}\\
        \midrule
        LSL & 25.8\% \\
        LSR & 24.88\% \\
        RSL & 24.6\% \\
        RSR & 24.68\% \\
        \bottomrule
    \end{tabular}
    \caption{The distribution of time-optimal path types based on Monte-Carlo simulations for BSB trajectories.}
    \label{table:distribution1}
\end{table}
\begin{table}[th]
        \centering
        \begin{tabular}{cc}
            \toprule
            \textbf{Decision table block} & \textbf{Distribution}\\
            \midrule
            $a_{11}$ & 5.26\% \\
            $a_{12}$ & 5.48\% \\
            $a_{13}$ & 5.46\% \\
            $a_{14}$ & 4.96\% \\
            $a_{21}$ & 5.51\%\\
            $a_{22}$ & 5.40\% \\
            $a_{23}$ & 5.44\% \\
            $a_{24}$ & 5.30\% \\
            $a_{31}$ & 5.99\% \\
            $a_{32}$ & 5.37\% \\
            $a_{33}$ & 5.78\% \\
            $a_{34}$ & 5.72\% \\
            $a_{41}$ & 5.54\% \\
            $a_{42}$ & 5.79\% \\
            $a_{43}$ & 5.16\% \\
            $a_{44}$ & 5.32\%\\
            \bottomrule
        \end{tabular}
        \caption{The distribution of Dubins decision table blocks from the Monte-Carlo simulations.}
        \label{table:distribution}
        \vspace{-3mm}
    \end{table}

The C++ package for our method is found at \href{https://github.com/castacks/trochoids}{https://github.com/castacks/trochoids}. The README contains instructions for building, running, and testing the codebase. Since the target application for this planner is UAVs, we also wanted the codebase to be able to handle small changes in altitude between start and goal states. To do this, we added a linear interpolation of the altitude between the start and goals states for all points in the path. For large altitude changes, future work could explore planning trochoidal paths in 3D, building upon the 3D Dubins planning of \cite{Owen2015}.

\section{Testing Results}\label{sec:results}

We validated our proposed method by performing Monte Carlo simulations. The start and goal poses $x_s^i$, $y_s^i$, $x_g^i$, and $y_g^i$ were randomly sampled from a uniform distribution in the range $[-1000, 1000]$ m. Similarly, $\psi_s^i$, $\psi_g^i$, and $\psi_w^i$ were uniformly sampled in the range [0, $2\pi$) rad. Additionally, the wind speed $V_w$ was randomly sampled in the range $[1, 15]$ m/s, and the turning radius $R_0$ was sampled in the range $[10,1000]$ m. The UAV airspeed was set as $V_a = 20$ m/s throughout all tests. Fig.~\ref{fig:three graphs} shows six randomly sampled trajectories.



The average run-time over $10,000$ sampled parameters for BSB trajectories on a 3.40 GHz CPU was $1.4224$ ms with a standard deviation of $8.006 \times 10^{-5}$ ms as compared to the baseline method, where two analytical methods and two numerical methods were iterated over, and the minimum-time path was selected \cite{techy2009}. This baseline method had an average solve time of $2.2711$ ms with a standard deviation of $2.97077 \times 10^{-5}$ ms. The testing results yielded a 37.4\% improvement over the baseline method, and the P value for the single tail difference between the two means is well less than $0.0001$. We found that 91.73\% of the 10,000 cases satisfy the $d > 4R_0$ condition mentioned in Section \ref{sec:method}, which determines the optimal trajectories are of type BSB.

\begin{figure}
     \centering
     \begin{subfigure}[b]{0.49\columnwidth}
         \centering
         \includegraphics[trim={0cm 0cm .9cm .9cm},clip,width=\textwidth]{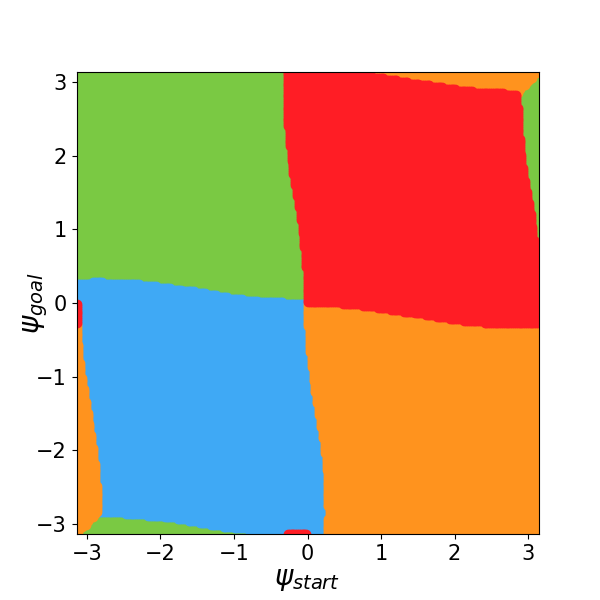}
         \caption{}
         \label{fig:psi_dubins}
     \end{subfigure}
     \begin{subfigure}[b]{0.49\columnwidth}
         \centering
         \includegraphics[trim={0cm 0cm .9cm .9cm},clip,width=\textwidth]{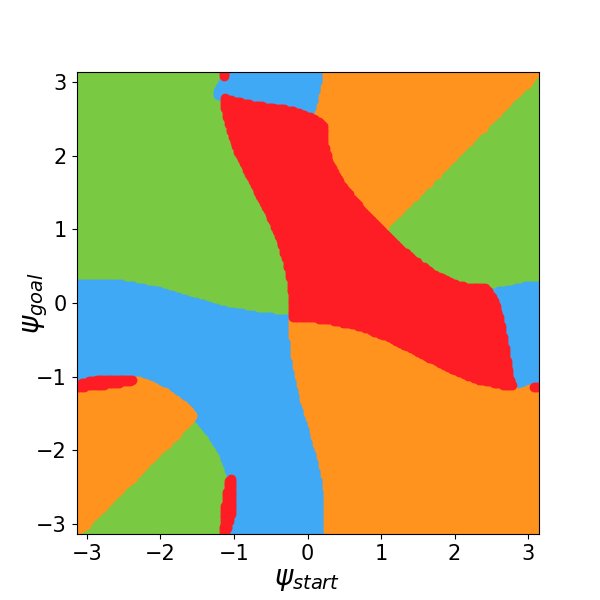}
         \caption{}
         \label{fig:psi_wind}
     \end{subfigure}
     \hfill
     \begin{subfigure}[b]{0.48\columnwidth}
         \centering
         \includegraphics[trim={0cm 0cm .9cm .9cm},clip,width=\textwidth]{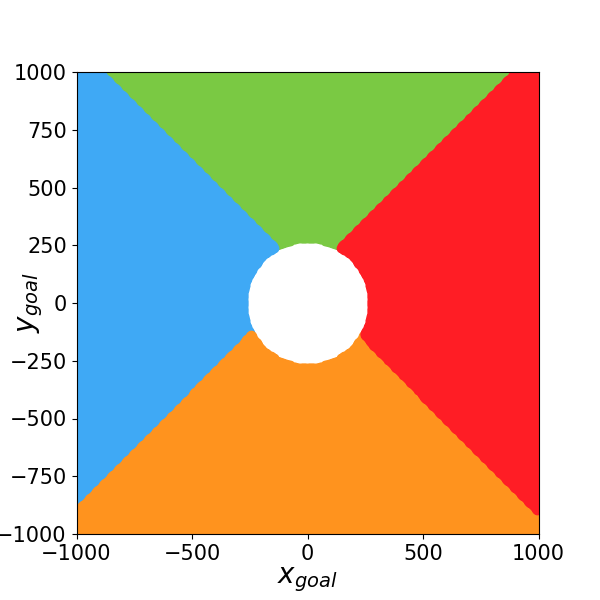}
         \caption{}
         \label{fig:states_dubins}
     \end{subfigure}
     \begin{subfigure}[b]{0.48\columnwidth}
         \centering
         \includegraphics[trim={0cm 0cm .9cm .9cm},clip,width=\textwidth]{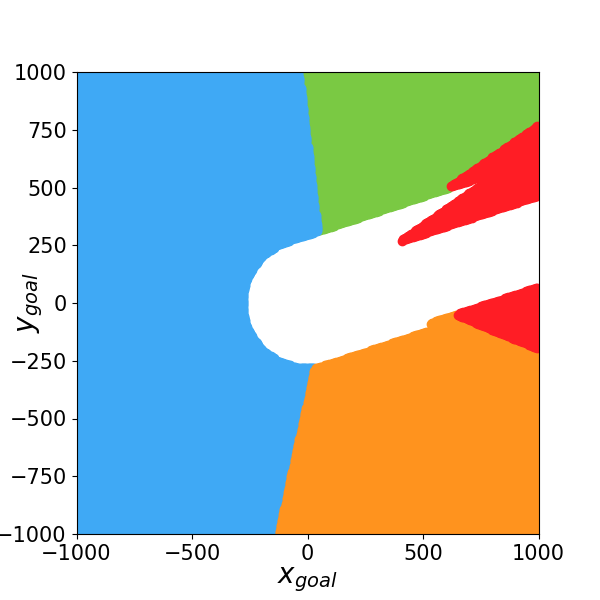}
         \caption{}
         \label{fig:states_wind}
     \end{subfigure}
        \caption{The first two figures show the distribution of solutions across varying $\psi_s^i$ and $\psi_g^i$ values and illustrate the difference between cases with wind speed zero (\ref{fig:psi_dubins}) and with wind speed $12.65$ m/s (\ref{fig:psi_wind}). The last two figures show the distribution of the solutions across varying start and goal states with wind speed zero (\ref{fig:states_dubins}) and wind speed $12.65$ m/s (\ref{fig:states_wind}). \textcolor{myRed}{RSL} is shown as red, \textcolor{myOrange}{RSR} as orange, \textcolor{myGreen}{LSL} as green, and \textcolor{myBlue}{LSR} as blue.}
        \label{fig:four graphs}
\end{figure}


Table \ref{table:distribution1} shows the distribution of the BSB trajectories for the $10,000$ runs, and the optimal candidate trajectory types are evenly spread out at approximately $25\%$. The distribution for the quadrant pairs of the tests within the Dubins decision table can be seen in Table \ref{table:distribution}. The overall distribution of trajectories within a specific block for the $10,000$ randomly sampled cases was also somewhat equivalent, showing full coverage of the testing domain.

We also visualize and compare the distributions of the optimal trajectory types for Dubins and trochoidal paths over various conditions in Fig.~\ref{fig:four graphs}. For these figures, $V_a = 20$ m/s, $V_w = 12.65$ m/s, $\psi_w = 0.322$ rad, and $R_0 = 70$ m. Fig.~\ref{fig:four graphs}\subref{fig:psi_dubins} shows the optimal Dubins trajectories, where the wind is ignored, for $x_s^i=0$, $y_s^i=0$, $x_g^i=470$ and $y_g^i=0$, and the initial and goal angles $\psi_s^i$ and $\psi_g^i$ are sampled from [$-\pi, \pi$]. The distribution of the trajectory types for wind cases can be seen in Fig.~\ref{fig:psi_wind}, depicting a shift in the decision boundaries for the wind cases. Similarly, start and goal angles were held constant at $\psi_s^i = {\pi}/{4}$ and $\psi_g^i = {3\pi}/{4}$ and the start and goal positions were sampled from a range of $[-1000,1000]$ m with the same $R_0$ and $V_w$ as mentioned above. The distribution for Dubins trajectories (ignoring the wind) can be seen in Fig.~\ref{fig:states_dubins}, whereas the distribution of trochoidal trajectories can be seen in Fig.~\ref{fig:states_wind}. The white hole in the center indicates states that do not satisfy the $d>4R_0$ condition and thus have not been included in the plot. The shift in trajectory type distributions from the effects of wind shows why the same switching functions from \cite{shkel2001} cannot be used directly for choosing the optimal trajectories.



To validate the robustness of this approach, we tested 5 million randomly sampled start states, goal states, wind speeds, and turning radii and compared the trajectory outputs of our method to exhaustively solving for all path types and choosing the best. Through this rigorous testing, we found the aforementioned error in the Dubins decision table, which occurred in only $3$ out of a million test cases. Fixing this error necessitated adding an additional path type to four quadrant pair cells, which correlated to additional numerical global root-finding operations (RSL and LSR solutions must be found numerically). Adding these extra operations affected the bounds of the maximum time improvements of our method. With the baseline having two numerical global root-finding problems for every solution, we could initially reduce the average number of numerical global root-finding problems to $1$ using the original decision table, achieving approximately a $50\%$ improvement from the baseline. However, to make this approach more robust, fixing the error in the decision table caused the average number of numerical global root-finding problems to increase to $1.25$, giving a maximum theoretical speed reduction of $37.4\%$.



\section{Conclusion} \label{sec:conclusion}
Based on the extensive testing results, the proposed method to compute time-optimal paths for curvature-constrained vehicles in a uniform wind outperforms baseline methods. The proposed approach reduces computation time, which is particularly important for real-time autonomous navigation systems such as UAVs, AUVs, or ASVs, especially in scenarios with limited computational capacities onboard. The future direction of this work is to explore fully implementing the switching functions for each quadrant block in the decision table rather than iterating over the potential trajectories in each cell, computing the lengths of the trajectories and choosing the time-optimal one. 



\section*{Acknowledgment}

We would like to thank Oswaldo Ramirez for his help on the video and proofreads for this work. 







\bibliography{ref}

\begin{figure}[t]
    \centering
    \includegraphics[trim={0cm 0cm 0cm .6cm},clip,width=.95\columnwidth]{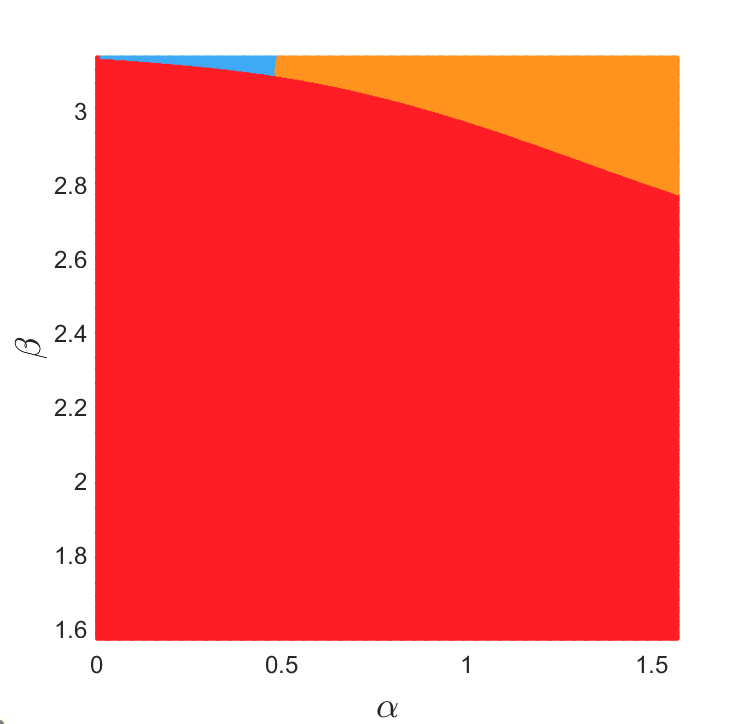} 
    \caption{~The shortest path distribution for case $a_{12}$ with $\alpha \in [0, \pi/2]$ and $\beta \in [\pi/2, \pi]$ with $d = 4.01R_0$. The red region marks the quadrant pairs for which the shortest path is \textcolor{myRed}{RSL}. The orange region marks the quadrant pairs for which the shortest path is \textcolor{myOrange}{RSR}, and the blue region marks the quadrant pairs for which the shortest path is \textcolor{myBlue}{LSR}.}
    \label{fig:a12_illustration}
\end{figure}
\appendix


In \cite{shkel2001}, a decision table is proposed to compute the shortest feasible path among the Dubins set $D=\{ LSL, RSR, RSL, LSR, RLR, LRL \}$, given the quadrants of $\alpha$ and $\beta$. After implementing and validating each block within this table, 4 cases were found, $a_{lm} \in \{ a_{12}, a_{21}, a_{34}, a_{43} \}$, where the previously proposed table did not give the actual shortest path option. Symmetrically, these four cases are equivalent. As such, the specific case of block $a_{12}$, which starts in quadrant 1 and finishes in quadrant 2, is addressed below.

According to Proposition 9 in \cite{shkel2001}, the shortest path for $a_{12}$ is either RSR or RSL based on the results of the switching function $S_{12}$. However, after comprehensive testing, there are cases where the shortest path is LSR. This error is particularly evident when the departing angle $\alpha$ is close to $0$ and the arriving angle $\beta$ is close to $\pi$. This was overlooked in the proof of Proposition 9 based on the differential analysis in the neighborhood of the edge case.

To visualize the solution space, we sampled across $\alpha$ and $\beta$ for $a_{12}$ with $d = 4.01R_0$ (to ensure no short path cases) and Fig.~\ref{fig:a12_illustration} shows the results of the shortest path distribution. The shortest path for all the quadrant pairs in the blue region is LSR, contradicting Proposition 9 in \cite{shkel2001}.

\begin{figure}[t]
    \centering
    \includegraphics[trim={0cm 0cm 0cm 0cm},clip,width=.48\textwidth]{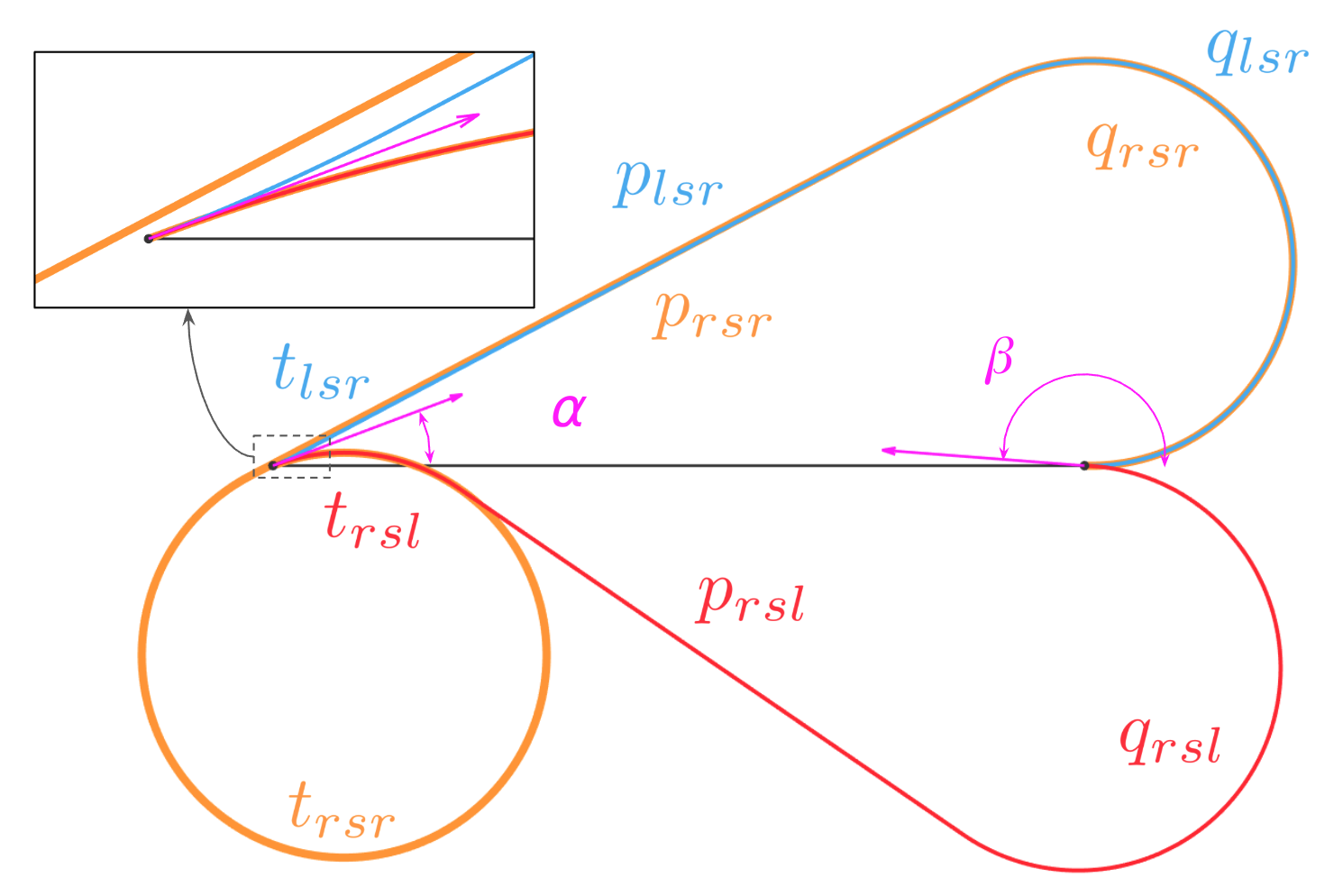} 
    \caption{~Plot of RSL, LSR, and RSR path for a point $\alpha = 0.36$ and $\beta = 3.111$ in the blue region of \ref{fig:a12_illustration}. Though the length of \textcolor{myBlue}{LSR} (blue) and \textcolor{myRed}{RSL} (red) are very close, the \textcolor{myBlue}{LSR} path is the shortest among all options after calculation. The area near the starting point is zoomed in to illustration the path difference better.}
    \label{fig:path_example}
\end{figure}

These counterexamples prove that the switching function presented in \cite{shkel2001}
$$S_{12} = p_{rsr} - p_{rsl} - 2(q_{rsl} - \pi)$$
sometimes does not select the actual shorter path between RSR and RSL. The $p_{rsr}$, $p_{rsl}$ and $q_{rsl}$ lengths of different sections of the path and are defined in \cite{shkel2001} and shown in Fig.~\ref{fig:path_example}, with the subscript referring to the trajectory type. For an example case shown in Fig.~\ref{fig:path_example}, the above switch function provided in \cite{shkel2001} would select RSR as the shortest path, but with nearly a whole circle as the first section of its path, RSR is obviously longer than the RSL path.

For our methodology, the decision table is modified so that it can still provide the shortest path for the cases shown above. The modified decision table is provided in Table \ref{table:new_matrix} with updated entries in $a_{lm} \in \{ a_{12}, a_{21}, a_{34}, a_{43} \}$ and was verified with over $5$ million random unit tests.

For further analysis of this issue, the corrected long path conditions, and the corrected switching functions in the decision table, please refer to \cite{Lim2023dubins}.

\end{document}